\documentclass[10pt,a4paper]{article}
\usepackage[utf8]{inputenc}
\usepackage[T1]{fontenc}
\usepackage{lmodern}
\usepackage[a4paper,margin=0.78in,headheight=28pt,headsep=10pt]{geometry}
\usepackage{microtype}
\usepackage{graphicx}
\usepackage{booktabs}
\usepackage{tabularx}
\usepackage{array}
\usepackage{amsmath,amssymb}
\usepackage{enumitem}
\usepackage{fancyhdr}
\usepackage{caption}
\usepackage{xcolor}
\usepackage{hyperref}
\usepackage{url}
\usepackage{titlesec}
\usepackage{float}

\definecolor{agblue}{RGB}{27,93,224}
\definecolor{aggray}{RGB}{105,105,105}
\hypersetup{
  colorlinks=true,
  linkcolor=agblue,
  citecolor=agblue,
  urlcolor=agblue,
  pdftitle={Learning-Based Speed Estimation from Accelerometer-Only Inertial Sensing},
  pdfauthor={Barak Or}
}

\setlist[itemize]{leftmargin=1.5em,topsep=0.25em,itemsep=0.15em}

\titleformat{\section}{\Large\bfseries\color{agblue}}{\thesection}{0.6em}{}
\titleformat{\subsection}{\large\bfseries\color{agblue}}{\thesubsection}{0.6em}{}
\titlespacing*{\section}{0pt}{1.35em}{0.55em}
\titlespacing*{\subsection}{0pt}{1.0em}{0.4em}
\newcommand{\keywords}[1]{\noindent\textbf{\color{agblue}Keywords.} #1}

\title{\vspace{-1.2cm}
{\small\color{aggray}\textbf{ArtificialGate Ltd. Research Preprint}}\\[0.55em]
\textbf{Learning-Based Speed Estimation from Accelerometer-Only Inertial Sensing}}
\author{
Barak Or\\
\small ArtificialGate Ltd., Israel\\
\small \href{mailto:barak@artificialgate.ai}{barak@artificialgate.ai}
\thanks{This research was conducted under ArtificialGate Ltd.}
}
\date{}

\begin{document}
\maketitle
\thispagestyle{fancy}

\begin{abstract}
The proposed model, CarSpeedNet, estimates scalar vehicle speed from a window of three-axis smartphone acceleration, without gyroscope, wheel-odometry, vehicle-bus, or positioning input at inference. The reported experiment comprises 13.2 hours of on-road driving. Beyond the network comparison, a finite-context analysis treats window length as part of the sensing problem. For nested histories, the minimum Bayes mean-square error is non-increasing with context; a complementary cue-coverage relation links the same window to the amount of speed-dependent vibration presented to the network and to the observation horizon. At a 1-s input, CarSpeedNet is compared with five temporal-network alternatives; the effect of temporal context is then measured over six window lengths. On the 0.5-hour holdout, a 4-s window yielded a root-mean-square error (RMSE) of 1.8 m/s and a mean absolute error (MAE) of 0.72 m/s, compared with 2.9 and 1.3 m/s at 1 s. The 178,169-parameter network requires about 0.68 MiB for 32-bit weights.
\end{abstract}

\keywords{accelerometer; intelligent sensors; inertial sensing; smartphone; vehicle speed estimation.}

\section{Introduction}
Wheel encoders, vehicle networks, and satellite navigation are the usual sources of vehicle speed. They are not always accessible to an auxiliary or low-cost sensing system, and direct integration of smartphone acceleration is quickly corrupted by bias, mounting orientation, and vibration \cite{farrell,mantouka}. Learning-based inertial methods offer a different route: they infer the state from temporal structure rather than integrating each sample \cite{brossard,liimu,oruncert,tong}.

Prior smartphone studies have combined acceleration with stop-based recalibration \cite{ustun}, used accelerometer--gyroscope pairs with recurrent networks \cite{deepspeed,abdelgawad}, or addressed phone placement and orientation through alignment, augmentation, and attention \cite{freydin_align,xiao,shin}. Our earlier speed estimator \cite{freydin_speed} used a six-axis inertial measurement unit (IMU), about 3 hours of data, RTK-GNSS supervision, and a dead-reckoning evaluation. The present experiment is distinct: inference uses only the three accelerometer axes, the reported corpus is 13.2 hours, five alternative temporal models are analyzed at 1 s, and CarSpeedNet is tested at six input lengths. Vehicle-vibration studies also indicate that inertial signatures can carry route- and motion-dependent information \cite{orvibration}.

\begin{figure}[H]
\centering
\includegraphics[width=0.58\textwidth]{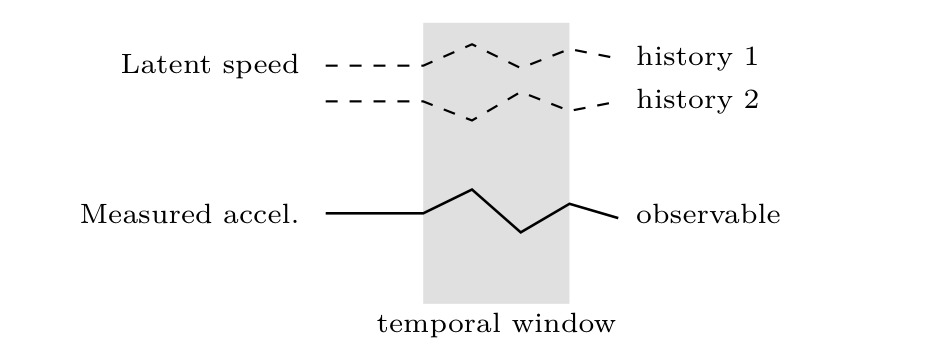}
\caption{Partial observability in accelerometer-only sensing. Distinct latent speed histories can be consistent with similar local acceleration. The shaded region marks the temporal context supplied to the estimator.}
\label{fig:partial}
\end{figure}

The main contribution is a finite-context characterization of the acceleration window supplied to CarSpeedNet in~\eqref{eq:mapping} and Figure~\ref{fig:partial}. For nested histories, the minimum Bayes mean-square error is non-increasing with context. We then connect the physical scale of a vibration cue to the finite input seen by the network by deriving how many cycles of that cue fall inside a $W$-sample window. This yields a minimum-context condition linking vehicle speed, spatial scale, sampling rate, and the observation horizon carried by each prediction. Across the six tested contexts, holdout RMSE decreases from 4.9 to 1.8 m/s and MAE from 2.50 to 0.72 m/s; the separate 1-s study compares CarSpeedNet with five temporal alternatives.

The remainder of the paper is organized as follows. Section~2 gives the sensing model and finite-context analysis, Section~3 describes the data and CarSpeedNet, Section~4 reports the holdout results, and Section~5 concludes.

\begin{table}[H]
\caption{Closest vehicle-speed studies and experimental scope.}
\label{tab:related}
\centering
\footnotesize
\begin{tabularx}{\textwidth}{p{0.16\textwidth} p{0.19\textwidth} p{0.25\textwidth} X}
\toprule
Study & Inference sensors & Estimation method & Reported scope \\
\midrule
Ustun and Cetin \cite{ustun} & Smartphone accelerometer & Integration with stop-based recalibration & Average error reported within 10 mph on its collected sample data \\
Freydin and Or \cite{freydin_speed} & Six-axis IMU & Recurrent DNN speed pseudomeasurement & Approximately 3 h; RTK-GNSS supervision; aided dead-reckoning evaluation \\
Shin et al. \cite{shin} & Smartphone accelerometer and gyroscope & Attention-LSTM regression & Four phones; 41.8 km; several orientations; OBD-II reference \\
Xiao et al. \cite{xiao} & Smartphone IMU & Pose alignment and sequence learning & Crowdsourced data; phone-placement augmentation; GNSS supervision \\
This paper & Smartphone accelerometer only & Recurrent--convolutional regression & 13.2 h; five alternatives at 1 s; six CarSpeedNet input lengths \\
\bottomrule
\end{tabularx}
\end{table}

\section{Accelerometer-Only Sensing}
A smartphone accelerometer measures specific force, not velocity. A simplified measurement model is
\begin{equation}
\mathbf{a}_m(t)=\mathbf{R}^{T}(t)\bigl(\dot{\mathbf{v}}(t)-\mathbf{g}\bigr)+\mathbf{b}(t)+\mathbf{n}(t),
\label{eq:accel}
\end{equation}
where $t$ denotes time, $\mathbf{a}_m(t)$ is the measured three-axis specific force, $\mathbf{R}(t)$ is the phone orientation rotation matrix, $\mathbf{v}(t)$ is vehicle velocity in the navigation frame, $\dot{\mathbf{v}}(t)$ is its time derivative, $\mathbf{g}$ is the gravity vector, $\mathbf{b}(t)$ is additive accelerometer bias, and $\mathbf{n}(t)$ is measurement noise. Short acceleration sequences are not sufficient to distinguish all velocity histories, particularly during constant-speed or low-dynamic motion. Figure~\ref{fig:partial} illustrates this partial observability. A longer window cannot make the problem fully observable, but it can expose stops, accelerations, and changes in vibration that are absent from an instantaneous sample.

CarSpeedNet implements the finite-horizon mapping
\begin{equation}
\hat{s}_t=f_{\boldsymbol{\theta}}\!\left(\mathbf{a}_{t-W+1:t}\right),
\label{eq:mapping}
\end{equation}
where $\hat{s}_t$ is the estimated scalar speed at time $t$, $f_{\boldsymbol{\theta}}$ is the learned regression function with parameters $\boldsymbol{\theta}$, $\mathbf{a}_{t-W+1:t}$ denotes the sequence of three-axis acceleration samples in the input window, and $W$ is the number of samples. The estimator learns correlations present in the training corpus; it does not separately identify orientation, bias, road excitation, or vehicle dynamics.

\subsection{Finite-Context Accuracy--Responsiveness}
Figure~\ref{fig:partial} makes the role of finite context explicit: the learned mapping in~\eqref{eq:mapping} receives only a $W$-sample segment of acceleration, while similar local measurements may remain compatible with different latent speed histories. Thus, $W$ controls the information presented to the network, not only the dimensions of its input. Let $\mathcal{A}_W=\mathbf{a}_{t-W+1:t}$ denote that history. Under squared-error loss, the Bayes-optimal estimate based on this window is $\hat{s}^{\star}_W=\mathbb{E}[s_t\mid\mathcal{A}_W]$. Its minimum mean-square error is
\begin{equation}
\mathcal{R}^{*}(W)=\mathbb{E}\!\left[(s_t-\hat{s}^{\star}_W)^2\right]=\mathbb{E}\!\left[\operatorname{Var}(s_t\mid\mathcal{A}_W)\right].
\label{eq:risk}
\end{equation}
The final term is the conditional variance averaged over the observed acceleration windows. If $W_2>W_1$, then $\mathcal{A}_{W_1}$ is contained in $\mathcal{A}_{W_2}$. Any estimator defined on the shorter history remains admissible on the longer history by ignoring the extra samples. Hence
\begin{equation}
\mathcal{R}^{*}(W_2)\leq\mathcal{R}^{*}(W_1),\qquad W_2>W_1.
\label{eq:monotonic}
\end{equation}
The monotonicity follows from the information set itself rather than from a particular architecture. Here, however, $\mathcal{A}_W$ is exactly the sequence presented to CarSpeedNet, so~\eqref{eq:monotonic} separates the information benefit of a longer input history from the choice of recurrent or convolutional layers.

Equation~\eqref{eq:monotonic} characterizes the information limit, but it does not show what additional temporal evidence becomes available to the network as $W$ grows. Every recurrent and convolutional layer in CarSpeedNet ultimately operates on the same finite $W$-sample record. To make the effect of that record explicit, let $x$ denote traveled distance and let $r(x)$ be one spatial vibration component with amplitude $A$ and wavelength $\lambda$:
\begin{equation}
r(x)=A\cos\!\left(\frac{2\pi x}{\lambda}\right).
\label{eq:spatial}
\end{equation}
Over a short interval with approximately constant speed $s$, $x(t)=x_0+st$. The component therefore reaches the accelerometer with temporal period
\begin{equation}
\tau(s,\lambda)=\frac{\lambda}{s},
\label{eq:period}
\end{equation}
equivalently at frequency $s/\lambda$. At sampling rate $f_s$, a $W$-sample input spans $T_W=W/f_s$ and contains
\begin{equation}
C_W(s,\lambda)=\frac{T_W}{\tau(s,\lambda)}=\frac{sW}{\lambda f_s}
\label{eq:coverage}
\end{equation}
cycles of that component. Thus, increasing $W$ changes a quantity directly available to the temporal network: the number of repetitions of a speed-dependent cue contained in one input. If at least $k$ cycles are to be present, the required context satisfies
\begin{equation}
W\geq W_k(s,\lambda)=\left\lceil\frac{k\lambda f_s}{s}\right\rceil.
\label{eq:mincontext}
\end{equation}
This relation couples the physical scale of the excitation to the network input length. It also exposes the responsiveness cost. The first full-context estimate requires $T_W=W/f_s$ seconds of measurements, and every later prediction still depends on that same span of history. Longer context therefore increases cue coverage and cannot worsen the Bayes limit, but it lengthens the observation horizon represented by each prediction. Section~4.2 examines this finite-context trade-off using the six available CarSpeedNet windows.

\section{Data and Model}
\subsection{Acquisition and Preprocessing}
The study reports 13.2 hours of highway and urban-road driving collected in Israel with a Samsung Galaxy smartphone. The stated allocation is 10.3 hours for training, 2.6 hours for validation, and 0.5 hours for holdout evaluation. The 0.5-hour subset is used as the holdout set.

Data collection began after stable GPS positioning had been confirmed from the geometric dilution of precision indicator \cite{sharp}. Acceleration was acquired at 500 Hz, low-pass filtered, and downsampled to 20 Hz. GPS speed was sampled at 1 Hz and supplied the training reference, but was not used at inference.

\begin{figure}[H]
\centering
\includegraphics[width=0.78\textwidth]{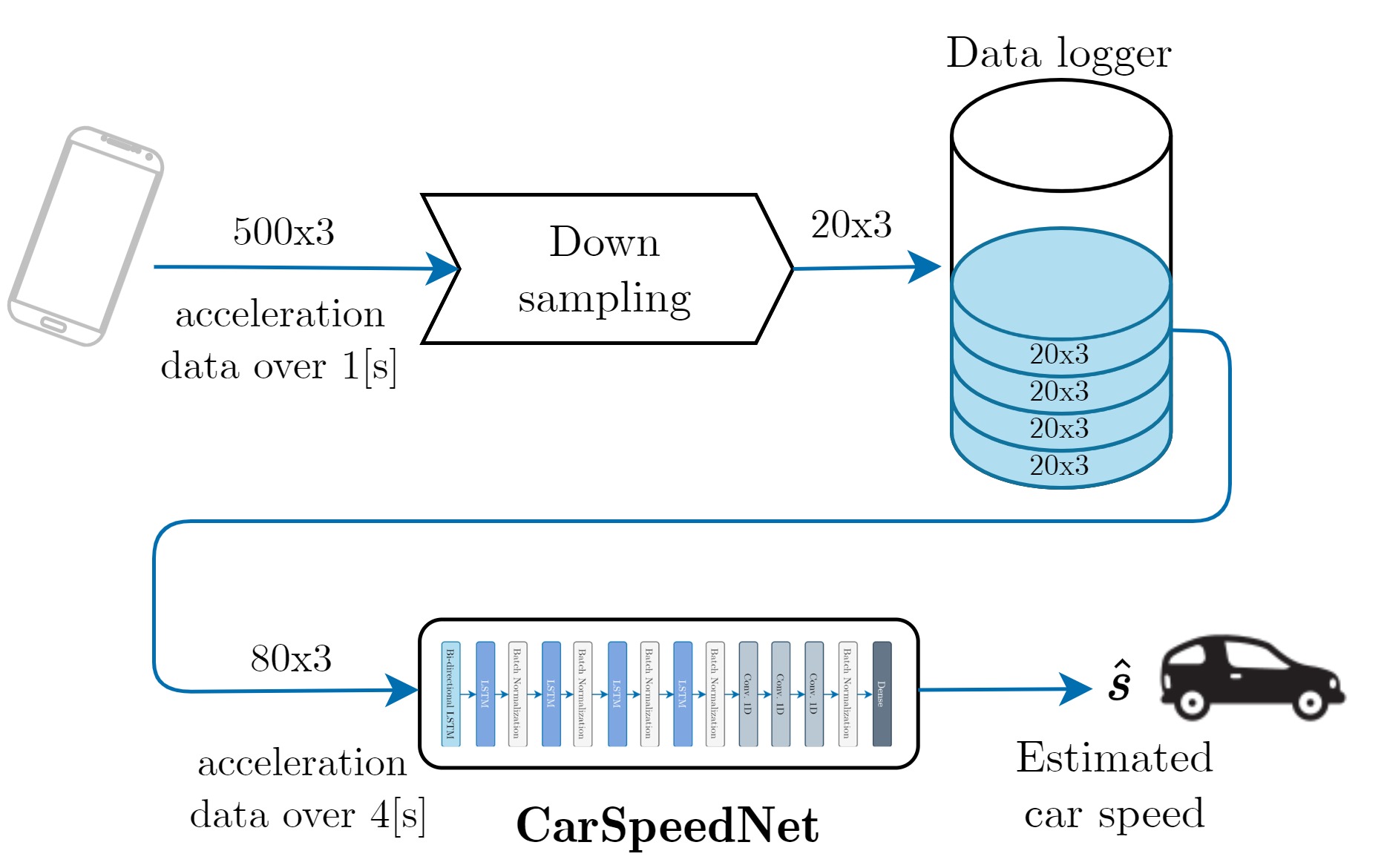}
\caption{Recorded sensing pipeline. One-second, 500-Hz acceleration blocks are downsampled to 20 samples; four blocks form the 80-sample CarSpeedNet input.}
\label{fig:pipeline}
\end{figure}

\subsection{CarSpeedNet}
\begin{figure}[H]
\centering
\includegraphics[width=0.74\textwidth]{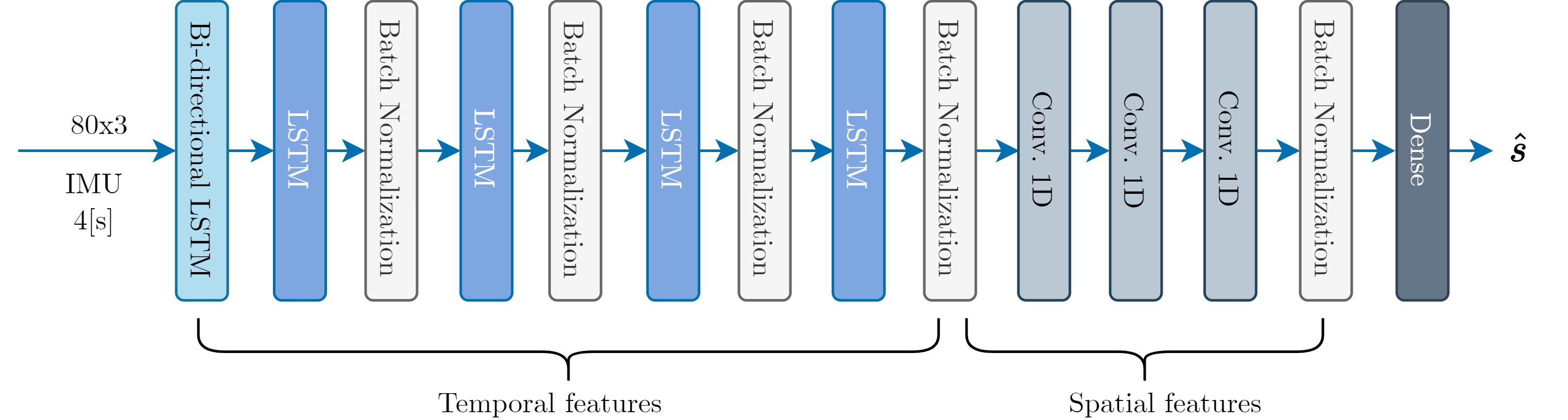}
\caption{Original CarSpeedNet architecture. The input contains 80 three-axis accelerometer samples (4 s at 20 Hz).}
\label{fig:arch}
\end{figure}

Figure~\ref{fig:arch} reproduces the original architecture. A 100-unit bidirectional LSTM is followed by four LSTM layers with 50, 20, 20, and 20 units and batch normalization. Three one-dimensional convolutional layers use 64, 64, and 32 filters with kernel size 3. A 32-unit fully connected layer and scalar output complete the model, with rectified-linear activations between applicable layers \cite{relu}. Training used Adam \cite{adam}, an initial learning rate of 0.001, batch size 32, and at most 200 epochs. The configuration lists a decay factor of 0.2 over 30,000 steps. CarSpeedNet has 178,169 trainable parameters, equivalent to 0.68 MiB of 32-bit weights, excluding activations and framework overhead.

\section{Holdout Results}
Performance is reported using root-mean-square error
\begin{equation}
\mathrm{RMSE}=\sqrt{\frac{1}{N}\sum_{j=1}^{N}\left(s^{\mathrm{GPS}}_j-\hat{s}_j\right)^2},
\label{eq:rmse}
\end{equation}
and mean absolute error
\begin{equation}
\mathrm{MAE}=\frac{1}{N}\sum_{j=1}^{N}\left|s^{\mathrm{GPS}}_j-\hat{s}_j\right|,
\label{eq:mae}
\end{equation}
where $N$ is the number of evaluated speed samples, $j$ is the sample index, $s^{\mathrm{GPS}}_j$ is the GPS-derived reference speed for sample $j$, and $\hat{s}_j$ is the corresponding CarSpeedNet estimate.

\subsection{Model Comparison}
\begin{figure}[H]
\centering
\includegraphics[width=0.72\textwidth]{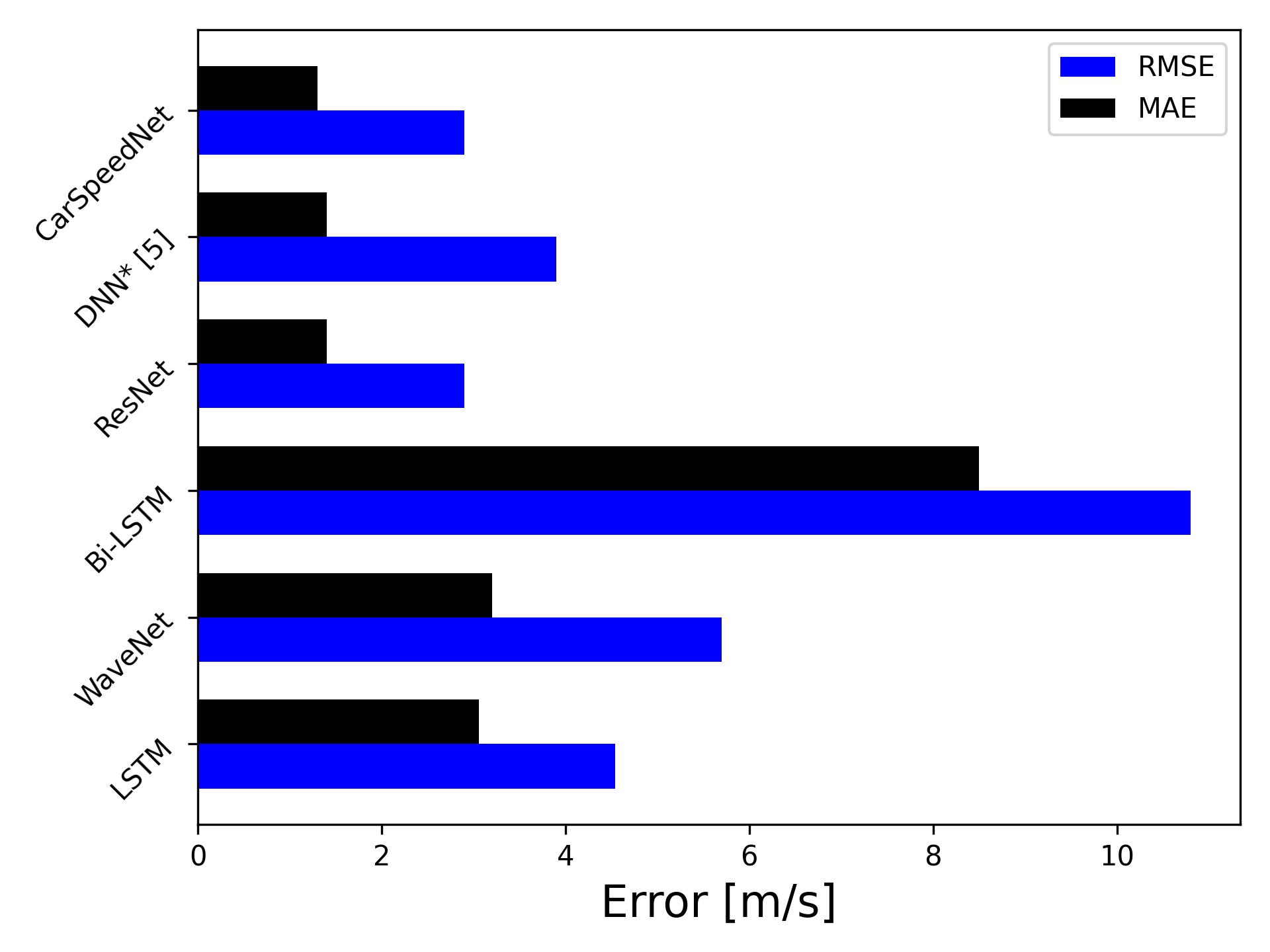}
\caption{Single-run holdout RMSE and MAE for six accelerometer-only models using a 1-s window.}
\label{fig:modelcomp}
\end{figure}

Five alternatives were trained on the same accelerometer-only input: DNN* adapted from \cite{freydin_speed}, a conventional LSTM \cite{lstm}, WaveNet \cite{wavenet}, a bidirectional LSTM, and a ResNet-inspired temporal model \cite{resnet}. Figure~\ref{fig:modelcomp} contains the single-run comparison for a 1-s window. CarSpeedNet yielded 2.9 m/s RMSE and 1.3 m/s MAE. The ResNet-inspired model had the same RMSE and 1.4 m/s MAE. Repeated-seed results and a common tuning budget are unavailable, so the small difference does not support a statistical ranking. CarSpeedNet is also not the smallest network: its 178,169 parameters are fewer than WaveNet's 239,937 but more than the other alternatives.

\subsection{Temporal Context}
Table~\ref{tab:window} provides the empirical counterpart to the risk relation in~\eqref{eq:monotonic}. RMSE decreases from 4.9 to 1.8 m/s as the window grows from 0.25 to 4 s; MAE decreases from 2.50 to 0.72 m/s. Both errors decrease monotonically across all six tested contexts, corresponding to reductions of 63.3\% in RMSE and 71.2\% in MAE between the shortest and longest windows. At 20 Hz, the tested inputs span observation horizons from 0.25 to 4 s. For any fixed speed $s$ and spatial scale $\lambda$,~\eqref{eq:coverage} therefore gives a 16-fold increase in cue coverage between the shortest and longest inputs, while the represented observation horizon also grows by the same factor.

\begin{table}[H]
\caption{Holdout accuracy across input windows.}
\label{tab:window}
\centering
\small
\begin{tabular}{ccc}
\toprule
Samples / duration & RMSE (m/s) & MAE (m/s) \\
\midrule
5 / 0.25 s & 4.9 & 2.50 \\
10 / 0.50 s & 4.2 & 2.10 \\
20 / 1.00 s & 2.9 & 1.30 \\
40 / 2.00 s & 2.3 & 1.00 \\
60 / 3.00 s & 2.2 & 0.88 \\
80 / 4.00 s & 1.8 & 0.72 \\
\bottomrule
\end{tabular}
\end{table}

\begin{figure}[H]
\centering
\includegraphics[width=0.82\textwidth]{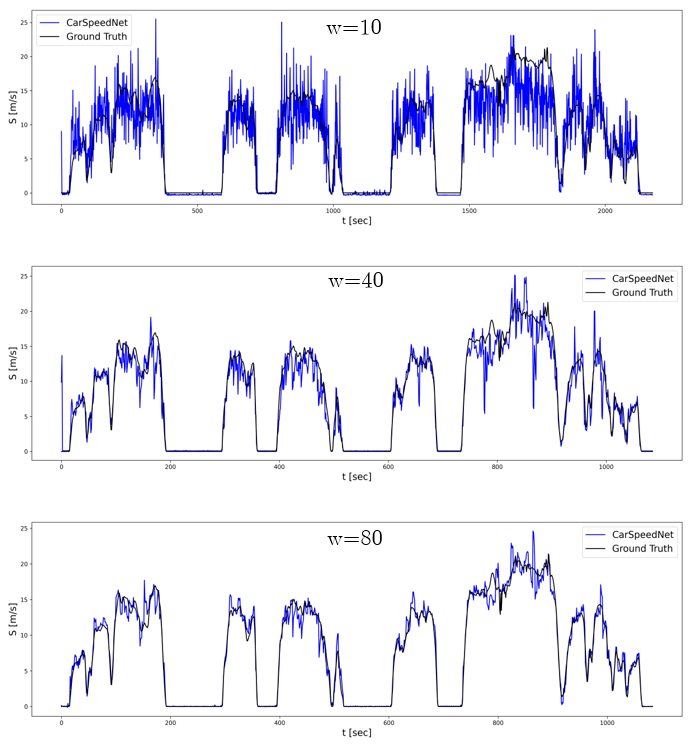}
\caption{Representative CarSpeedNet traces and 1-Hz GPS-derived reference for $W\in\{10,40,80\}$. The embedded legend retains the original label ``Ground Truth.''}
\label{fig:traces}
\end{figure}

Figure~\ref{fig:traces} reproduces the saved traces for $W\in\{10,40,80\}$. In the displayed intervals, the estimate approaches zero with the GPS reference and becomes smoother as the window grows. Quantitative comparisons rely on Table~\ref{tab:window}.

\subsection{Experimental Limitations}
The evaluation is based on the available recorded corpus and does not include a dedicated cross-device or cross-vehicle generalization study. Robustness to different smartphone placements, sensor characteristics, and driving environments remains to be evaluated in future work.

\section{Conclusion}
CarSpeedNet estimates vehicle speed from smartphone acceleration alone. The finite-context analysis treats window length as part of the sensing problem rather than only a network setting. A longer nested history cannot increase the minimum Bayes mean-square error. For speed-dependent vibration, the same window also determines how many repetitions of a physical spatial scale are exposed to the temporal network, while $T_W=W/f_s$ sets the history carried by each prediction. The main empirical result is the reduction from 4.9/2.50 m/s RMSE/MAE at 0.25 s to 1.8/0.72 m/s at 4 s on the recorded holdout. At 1 s, CarSpeedNet and the ResNet-inspired alternative have the same RMSE. The monotonic window-length trend is consistent with this finite-context view: longer windows provide a larger information set and more repeated temporal evidence, at the cost of a longer observation horizon.

\section*{Conflict of Interest}
The author declares no conflict of interest.

\end{document}